\pdfoutput=1
\documentclass[11pt]{article}

\usepackage[preprint]{acl}

\usepackage{makecell}

\usepackage{times}
\usepackage{latexsym}

\usepackage[T1]{fontenc}
\usepackage[utf8]{inputenc}

\usepackage{microtype}

\usepackage{inconsolata}

\usepackage{graphicx}
\usepackage{subfig}
\usepackage{amsmath,amsfonts,amssymb,amsthm}
\usepackage{bm}
\usepackage{booktabs}
\usepackage{xcolor}
\usepackage{colortbl}
\usepackage{stmaryrd}
\usepackage{url}
\usepackage{hyperref}
\usepackage{multirow}
\usepackage{verbatim}

\usepackage{amsmath}
\usepackage{amsmath,amsfonts,bm}

\def\eqref#1{equation~\ref{#1}}
\def\1{\bm{1}}

\def\vb{{\bm{b}}}
\def\vc{{\bm{c}}}

\def\vk{{\bm{k}}}

\def\vw{{\bm{w}}}
\def\vx{{\bm{x}}}

\def\evb{{b}}
\def\evc{{c}}

\def\evx{{x}}

\def\mQ{{\bm{Q}}}
\def\mR{{\bm{R}}}
\def\mS{{\bm{S}}}
\def\mT{{\bm{T}}}

\def\mV{{\bm{V}}}

\DeclareMathAlphabet{\mathsfit}{\encodingdefault}{\sfdefault}{m}{sl}
\SetMathAlphabet{\mathsfit}{bold}{\encodingdefault}{\sfdefault}{bx}{n}

\newcommand{\R}{\mathbb{R}}

\newcommand{\softmax}{\mathrm{softmax}}

\DeclareMathOperator*{\diag}{diag}
\DeclareMathOperator{\atantwo}{atan2}

\title{DocPolarBERT: A Pre-trained Model for Document Understanding\\with Relative Polar Coordinate Encoding of Layout Structures}

\author{Benno Uthayasooriyar\textsuperscript{{\normalfont 1, 2}} \qquad Antoine Ly\textsuperscript{{\normalfont 1}} \qquad Franck Vermet\textsuperscript{{\normalfont 2}} \qquad Caio Corro \textsuperscript{{\normalfont 3}}
\\
\textsuperscript{1}Data Analytics Solutions, SCOR
\quad
\textsuperscript{2}Univ Brest, CNRS, UMR 6205, LMBA
\\
\textsuperscript{3}INSA Rennes, IRISA, Inria, CNRS, Université de Rennes
}

\begin{document}
\maketitle
\begin{abstract}
We propose a novel self-attention mechanism for document understanding that takes into account text block positions in relative polar coordinate system rather than the Cartesian one.
Based on this mechanism, we build \textsc{DocPolarBERT}, a layout-aware \textsc{BERT} model for document understanding that eliminates the need for absolute 2D positional embeddings.
Despite being pre-trained on a dataset more than six times smaller than the widely used \textsc{IIT-CDIP} corpus, \textsc{DocPolarBERT} achieves state-of-the-art results.
These results demonstrate that a carefully designed attention mechanism can compensate for reduced pre-training data, offering an efficient and effective alternative for document understanding.
\end{abstract}

\section{Introduction}
The standard setting in natural language processing is to consider text that appears as sequential data.
In contrast, document understanding aims to analyze data that is presented in two dimensional documents whose layout structure convey crucial information, such as forms and invoices.
To this end, one can extend large language models (LLM) to incorporate layout information \cite{wang2023docllm, tanaka2024instructdoc,Luo_2024_layoutllm}.
However, LLMs pose challenges for real-world deployment due to their high computational costs, often causing bottlenecks when processing large volumes of data in real-time applications.
Furthermore, API-based models are often a \emph{no go} when data privacy is a critical concern.

As such, developing light and efficient models is a critical line of work.
Similar to \textsc{BERT} \cite{devlin_bert_2019},
prior work proposed to pre-train self-attentive networks \cite[i.e.\ transformer encoders][]{vaswani_attention} on large amount of unlabeled documents \cite{xu_layoutlm_2020,Huang2022LayoutLMv3PF, tu2023layoutmask, jiang-etal-2025-relayout}.
These models can then be fine-tuned for downstream tasks like named-entity recognition.

A key characteristic of any document \textsc{BERT} model is how they encode layout structures.
Previous works rely on absolute position encoding,
often with the addition of a relative position information encoded as bias in the attention mechanism \citep{xu_layoutlmv2_2022}, in order  to improve spatial interaction modeling. 
We hypothesize that such encoding of layout information can lead to generalization issue.
First, absolute positional embeddings are not invariant to translation, which is a key characteristic of many documents of interest: the representation of elements in a table should not be impacted by the position of the table in the document.
Second, relative positions between elements always rely on the Cartesian coordinate system.
This obfuscate key information (``is above of'', ``is at a given distance of'') and can hinder generalization due to data sparsity issues.

As such, we argue that the polar coordinate system can better leverage layout information.
Indeed, when dealing with relative positions in forms and tables, the polar coordinate system that distinguishes between angle and distance is a more semantically appropriate choice, capturing:
\begin{itemize}
    \item Angular information: column headers are often in first row, and this information must prevail for all other rows;
    \item Distance information: related fields in a form are close (e.g.\ first name, last name, birth date), but their order may differ.
\end{itemize}
Therefore, in this work we propose a novel \textsc{BERT}-like encoder for documents that completely removes any absolute positional information, and relies only on relative encoding of layout structures in the polar coordinate system.

Our contributions can be summarized as follows:
(1)~we propose a novel attention mechanism that incorporates relative positional encoding in the polar coordinate system;
(2)~we pre-train such a model on publicly available data;
(3)~we evaluate our model by fine-tuning on several downstream tasks, and show it achieves competitive or better results than comparable baselines;
and (4) we also conduct a comprehensive set of experiments and ablation studies to validate our design choices and assess the robustness of the proposed approach, covering generalization across complex layouts, discretization strategies of distances, fair comparison with a vison-free \textsc{LayoutLMv3} variant, attention pattern analysis, and computational efficiency.

Code, data and models are publicly available.\footnote{\url{https://github.com/buthaya/docpolarbert}}

\section{Related Work}

\paragraph{Document understanding.} 
\citet{xu_layoutlm_2020} introduced absolute positional embeddings that encode bounding boxes of each text unit (word or group of words) in the input of a \textsc{BERT} model.
Following in this direction, researchers have developed multi-modal encoders that include extra vision features, often leading to specific training strategies \cite{appalaraju2021docformer, gu2021unidoc, xu_layoutlmv2_2022,Huang2022LayoutLMv3PF, bai-etal-2023-wukong}. 
Nevertheless, as vision models are computationally intensive, there is an interest in alternatives without vision features, to achieve fast inference while maintaining competitive performance \cite{hong2022bros,wang-etal-2022-lilt,tu2023layoutmask,jiang-etal-2025-relayout}.

\paragraph{Relative attention.}
\citet{shaw_self-attention_2018} introduce \emph{relation-aware self-attention}, which extends the standard attention mechanism with biases based on relative positions of tokens.
This design embeds positional relationships directly into the attention process, rather than depending only on absolute position embeddings.
Recent research on document understanding \cite{tilt_2021, xu_layoutlmv2_2022, hong2022bros, Huang2022LayoutLMv3PF} follows similar strategies by adapting self-attention to handle two-dimensional spatial relationships.
In these models, the attention bias between two text bounding boxes is computed according to their relative spatial arrangement: (1) pairwise relative distances are computed following a specified geometric definition, i.e.\ distance between top-left corners of the boxes, (2) these continuous distances are discretized into fixed buckets and (3) each discrete value is mapped to a learned embedding used as a bias in the attention computation.
Closer to our work, \citet{hwang2021spatialdep} also used relative coordinates in the polar coordinate system.

\section{Neural Architecture}

\begin{figure*}
\centering
\subfloat[\textsc{LayoutLMv2/v3}]{\label{a}\includegraphics[width=.25\linewidth]{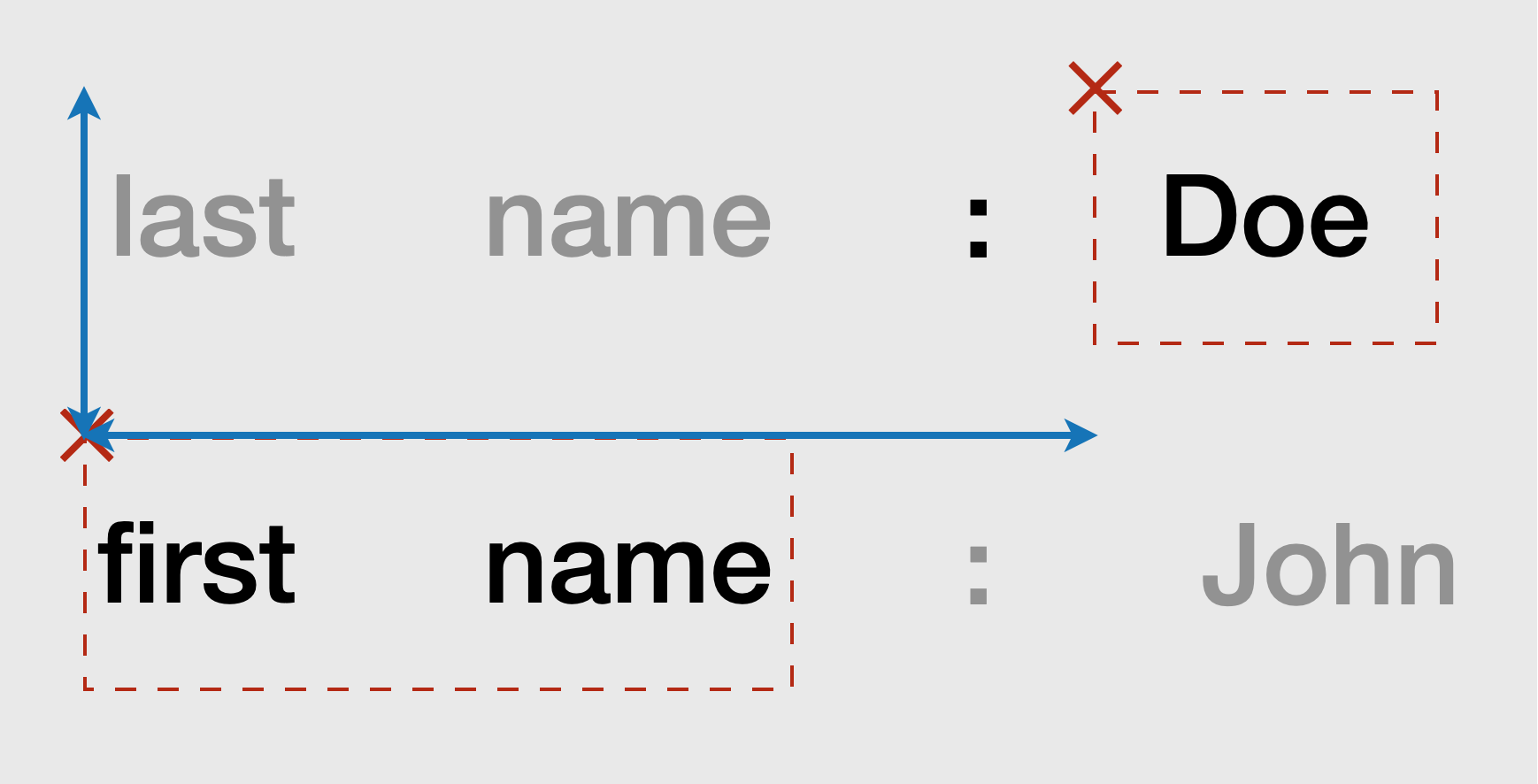}}\label{fig:spatial_bias_layoutlmv3}\hfill
\subfloat[\textsc{BROS}]{\label{b}\includegraphics[width=.25\linewidth]{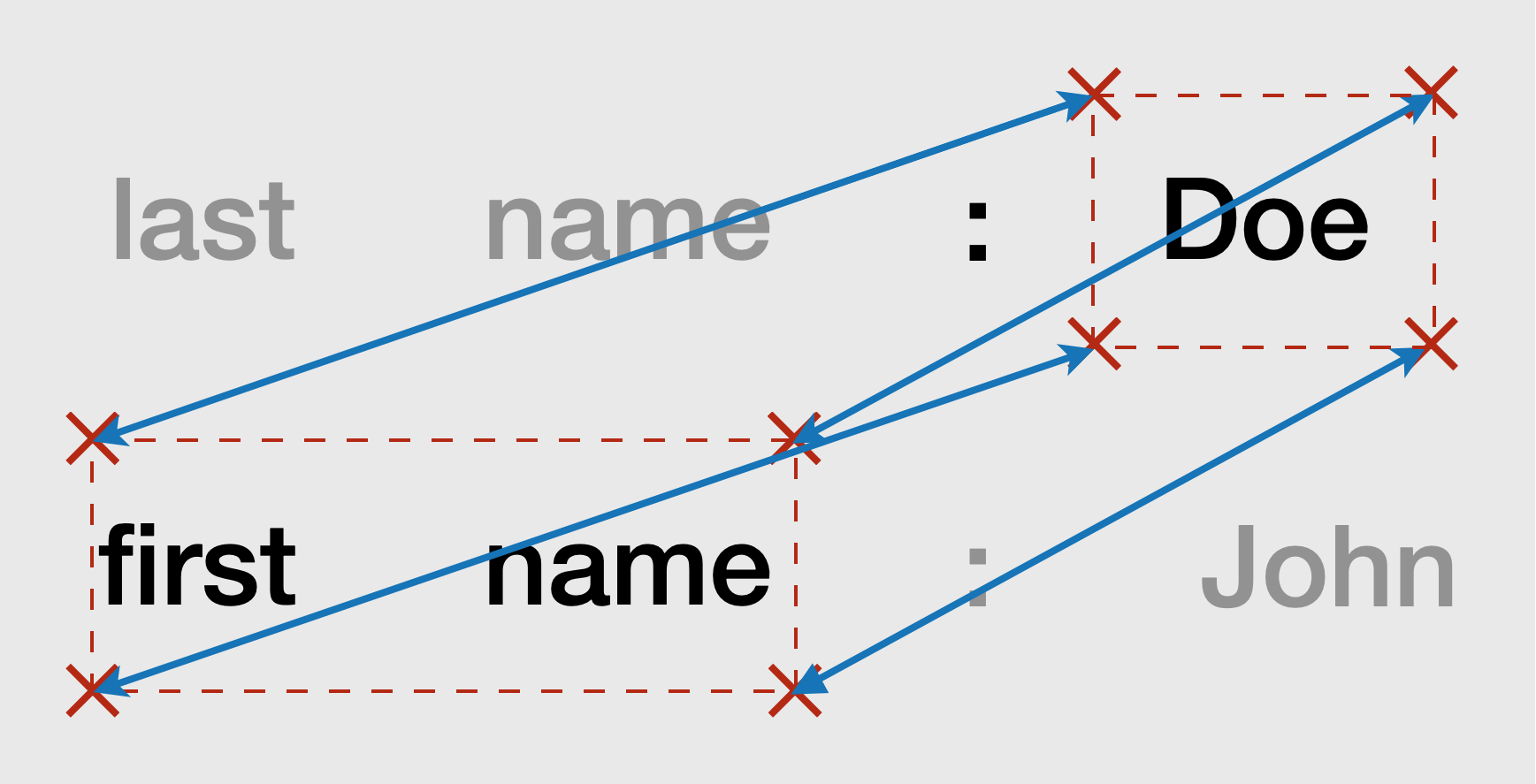}}\hfill
\subfloat[Our method]{\label{c}\includegraphics[width=.25\linewidth]{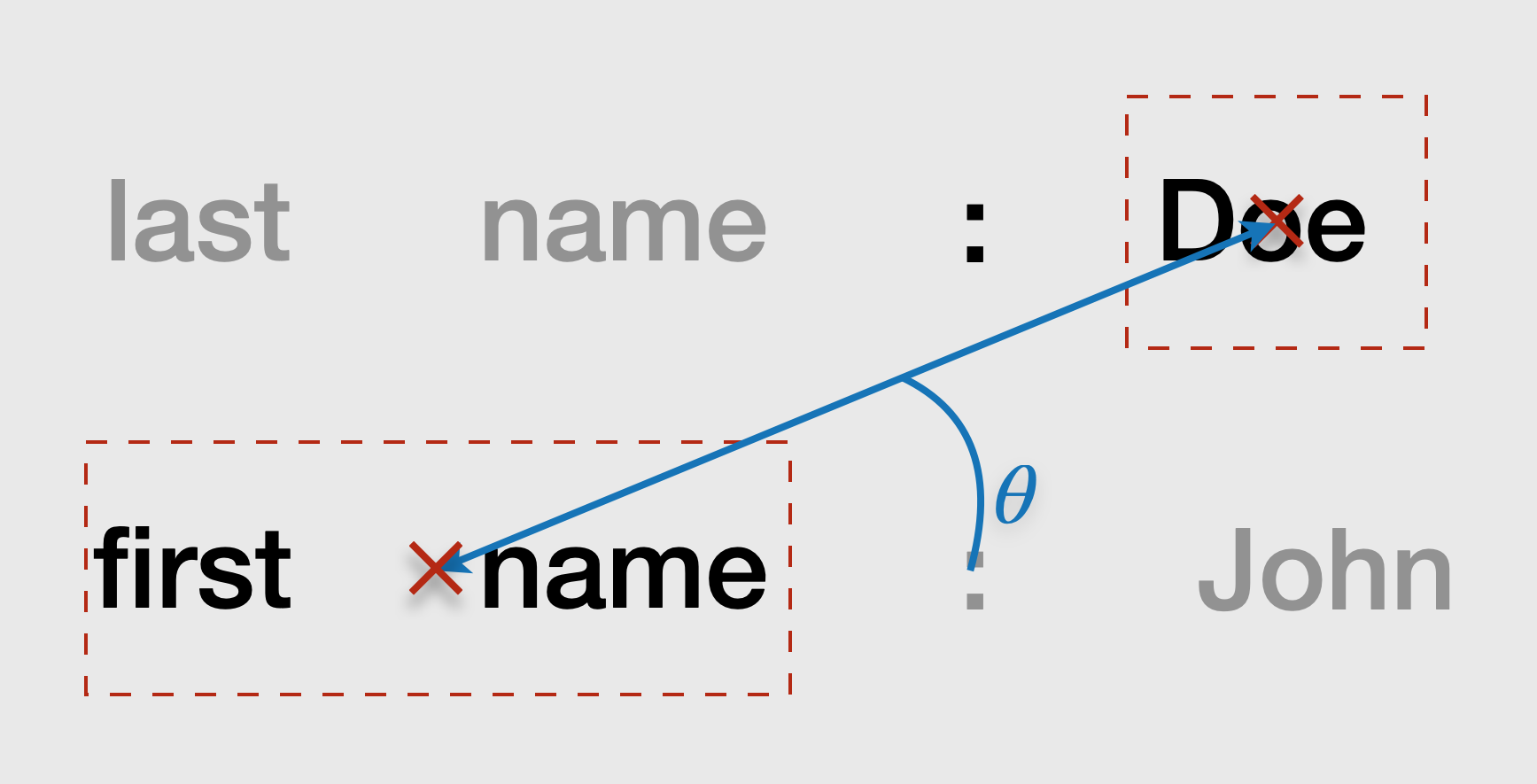}}
\vspace{-0.3cm}
\caption{Illustration of relative positional attention in different models.
Multiple words can share the same bounding box.
Attention is skewed with respect to:
\textbf{(a)} the horizontal and vertical distances between the top left corners;
\textbf{(b)} distance between the four corners, respectively;
\textbf{(c)} distance and angle between centers of the two bounding boxes.%
}
\label{fig:spatial_bias}
\end{figure*}

In this section, we describe our novel encoding method of layout structures using relative positional information in the polar coordinate system. 

\subsection{Background: Relative Attention}

Without loss of generality, in the following we assume inputs of $n$ tokens and that query, key and value vectors have the same dimension $d > 0$ to simplify notation.
We denote $\mQ, \mV \in \R^{n \times d}$ the query and value matrices,
and $\vk \in \R^d$ the (single) key vector.
For example, for a sentence of $n$ tokens,
we consider the attention mechanism for a single token, represented by key $\vk$,
and $\mQ_j$ and $\mV_j$ are the query and value vectors associated with input at position $j$, respectively.\footnote{We write $\mQ_i$ the $i$-th row of matrix $\mQ$ as a column vector.}

The attention mechanism computes an output vector $\vw \in \R^d$ as follows \cite{vaswani_attention}:
\[
\vw \triangleq \mV^\top \softmax\left(\sqrt{d}^{-1}\mQ\vk\right)\,.
\]
To introduce relative positional information,
\citet{shaw_self-attention_2018} augment the attention mechanism with input dependent modifications of the key.
Let $\mR \in \mathbb{R}^{n \times d}$ be a matrix where $\mR_j$ is a vector encoding the relative distance with the $j$-th input, e.g\ if $\vk$ is associated with token $i$, $\mR_j$ could be set to an embedding representing the distance $i - j$.
The attention mechanism is modified as follows:
\begin{align*}
\vw &\triangleq \mV^\top \softmax\left(
\sqrt{d}^{-1}
\begin{bmatrix}
    \mQ_1^\top(\vk + \mR_1) \\
    \vdots \\
    \mQ_n^\top(\vk + \mR_n)
\end{bmatrix}
\right), \\
\intertext{which can be conveniently rewritten as:}
&= \mV^\top \softmax\left(\sqrt{d}^{-1}\left(\mQ \vk + \diag(\mQ\mR^\top)\right)\right),
\end{align*}
where $\diag(\cdot)$ is the matrix diagonal as a vector.

\subsection{Input Representation}

In practice, inputs are documents that have been processed using an optical character recognition (OCR) model.
The output of the OCR model is a sequence of tokens $\vx = (\evx_1\dots\evx_n)$ and a sequence of bounding boxes $\vb = (\evb_1\dots\evb_n)$, where $\evb_i$ contains information about the position and size of the box containing the $i$-th token.
Several tokens may share the same bounding box information, i.e.\ the $i$-th and $(i+1)$-th tokens can be in the same box.
Note that even though tokens appear in the two-dimensional structure of the document,
the linearization provided by the OCR can give crucial information for downstream tasks,
and is therefore preserved in practice.

Tokens are embedded as follows:
\[
\texttt{E}(\evx_i) \triangleq \texttt{E}_{\text{voc}}(\evx_i) + \texttt{E}_{\text{1D-pos}}(i) + \underbrace{\texttt{E}_{\text{2D-pos}}(\evb_i)}_{\text{for ablation only}},
\vspace{-0.3cm}\]
where $\texttt{E}_{\text{voc}}(\cdot)$ is a standard token embedding table,
$\texttt{E}_{\text{1D-pos}}(\cdot)$ and $\texttt{E}_{\text{2D-pos}}(\cdot)$ are learned (absolute) positional embedding tables.
Following \citet{xu_layoutlm_2020},
1D positional embeddings are similar to the ones of \textsc{BERT} \cite{devlin_bert_2019},
whereas 2D positional embeddings are the sum of four learned embeddings representing discretized absolute position of the bounding box of the token.

\subsection{Relative Polar Coordinate Encoding}

Let $\vc_i = [\evc_{i,1}, \evc_{i,2}]^\top$ be the center of the bounding box $\evb_i$ containing the $i$-th token expressed in the Cartesian coordinate system.
Defining polar coordinates requires specifying an origin and a polar axis.
As we aim to represent relative position between one input token (query position)
and all other context tokens (key positions),
we choose the center of the bounding box including the considered input (query) token as the origin.
The polar axis is set to the first Cartesian dimension.

In the following, we assume the considered input is the $i$-th token.
The relative position of the $j$-th with respect to the $i$-th one in the polar coordinate system is the tuple $\left(\rho(\vc_i, \vc_j), \theta(\vc_i, \vc_j\right))$ with:
\begin{align*}
    \rho(\vc_i, \vc_j) &\triangleq \|\vc_i - \vc_j\|_2\,,\\
    \theta(\vc_i, \vc_j) &\triangleq \atantwo(\vc_j, \vc_i),
\end{align*}
where we extend the $\atantwo$ to be null when its two inputs are equal, i.e.\ $\atantwo(\vc_i, \vc_i) = 0$.
These relative polar coordinates are embedded using two learned embedding tables $\texttt{E}_{\text{dist.}}(\cdot)$ and $\texttt{E}_{\text{angle}}(\cdot)$.
To this end, we discretize the real values.
For distances, we cap the maximum to a constant $\rho_{max}$ and equally divide $[0, \rho_{max}]$ into 4 values.
For angles, we discretize $\left( -\pi/2, \pi/2 \right)$ into 8 values.
More information is given in Section~\ref{sec:docpolarbert_bucketing}.

Let $\vk$ be the key associated with the $i$-th input,
and $\mQ$ and $\mV$ defined as previously.
Our relative polar coordinate attention mechanism is defined as follows.
We build matrices $\mS, \mT \in \R^{n\times d}$ as:
\begin{align*}
    \mS_j &\triangleq \texttt{E}_{\text{dist.}}(~\rho(\vc_i, \vc_j)~), \\
    \mT_j &\triangleq \texttt{E}_{\text{angle}}(~\theta(\vc_i, \vc_j)~).
\end{align*}
Then, the output vector $\vw$ is defined as:
\[
\vw
\triangleq \mV^\top \softmax\hspace{-0.05cm}\left(\sqrt{d}^{-1}\hspace{-0.15cm}\left(
\begin{array}{l}\displaystyle
     \mQ \vk \\
     + \diag(\mQ\mS^\top) \\
     + \diag(\mQ\mT^\top) \\
\end{array}\right)\right)\hspace{-0.07cm}.
\]
This modified attention mechanism accounts for relative distances between tokens, as well as relative angles around them in the polar basis.

Figure~\ref{fig:spatial_bias} illustrates the key differences of our approach.
The full model is illustrated in App.~\ref{appendix:architecture}.

\section{Positional Embeddings \& Discretization}
\label{sec:discretization}

\begin{figure*}[t]
\centering
\subfloat[Distances $\rho$, discretized using four different buckets.]{\label{a}\includegraphics[width=0.45\textwidth]{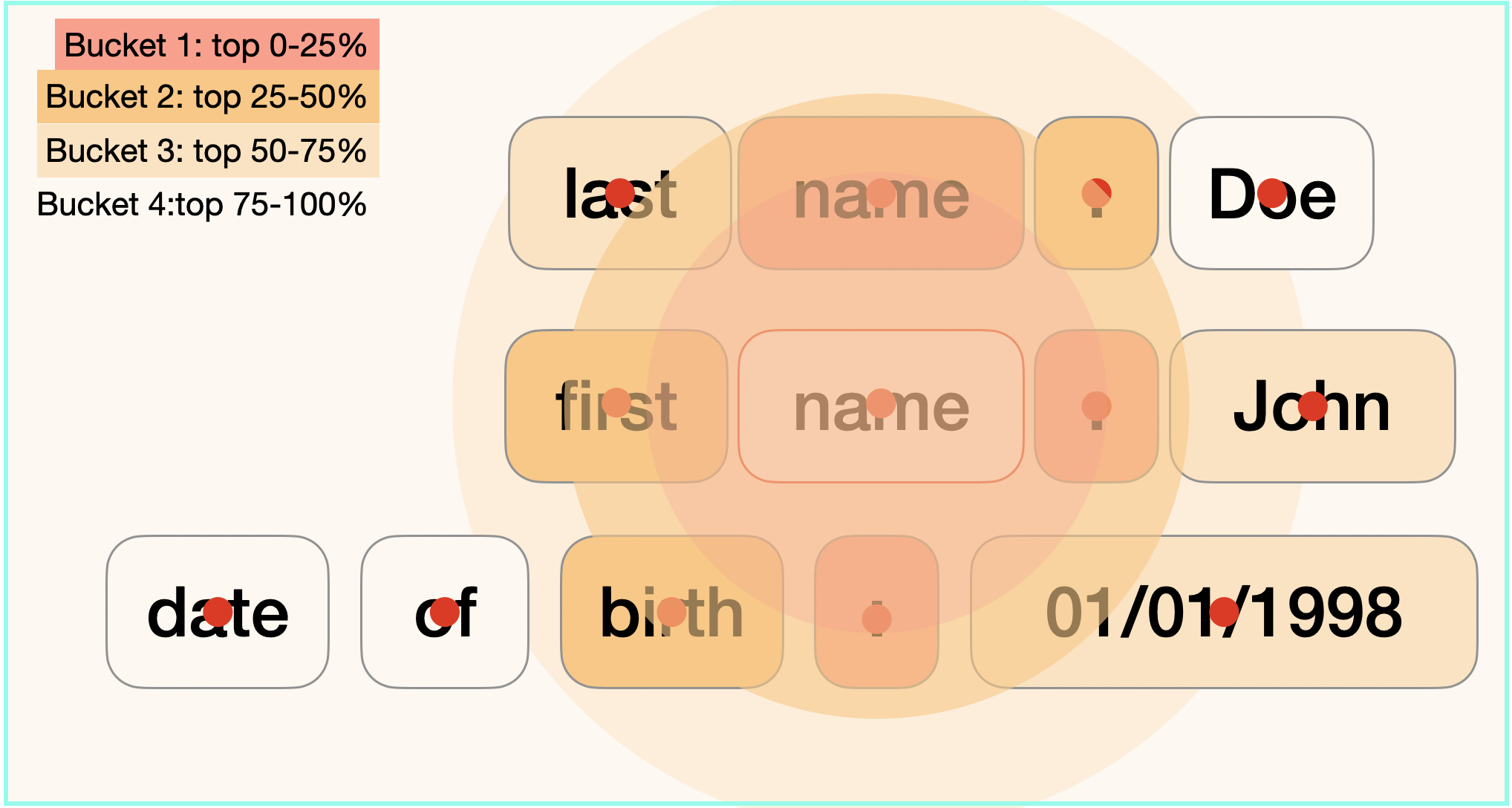}}
\hspace{1cm}
\subfloat[Angles $\theta$.]{\label{b}\includegraphics[width=0.45\textwidth]{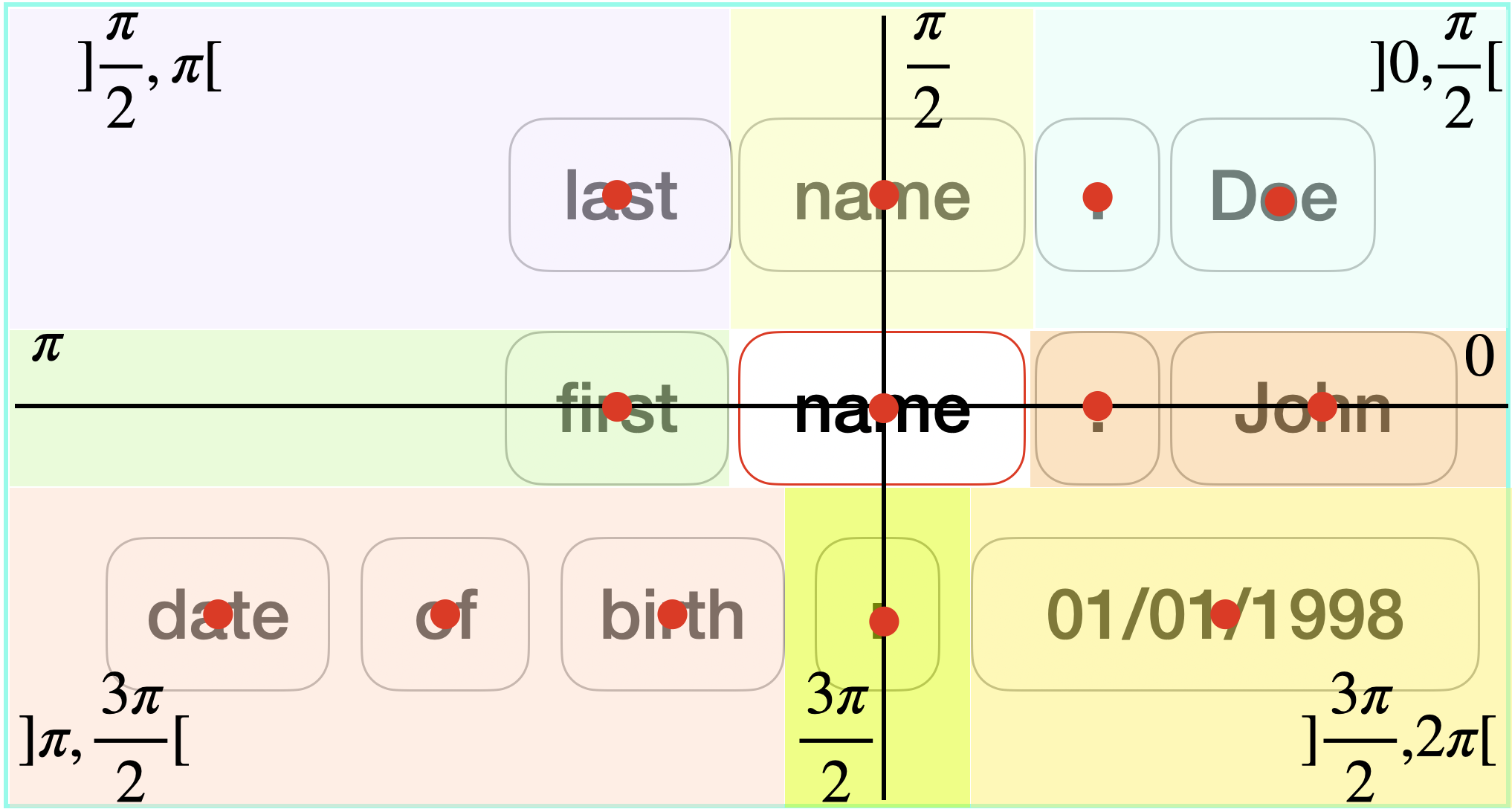}}
\caption{Illustration of relative positions for the token ``name'' and its neighbors in polar coordinates.}
\label{fig:discretization_methods}
\end{figure*}

Positional embeddings are little discussed in previous work.
In this section, we first review the main discretization strategies for absolute and relative positional embbeddings, and then we motivate and propose a new approach for relative distances.

\subsection{Absolute Positions in \cite{xu_layoutlm_2020}}

In \textsc{LayoutLM} \cite{xu_layoutlm_2020},
absolute token position embeddings are added to token embedding in the network input.
For each token $\evx_i$, the model adds
(1)~a \emph{textual position embedding} which corresponds to its sequence index $i$ (regardless of its spatial position)
and (2)~a \emph{layout embedding} corresponding to its bounding box $\evb_i$ position.

In order to embed the bounding box position, its continuous position information must first be discretized.
This transformation can be written as follows.
Let $c \in \R$ be one of the four ``raw'' coordinates of a bounding box $\evb_i$ (top-left, top-right, bottom-left or bottom-right).
Its discretized embedding index $c'$ is computed as follows:
\[
c' = \left\lfloor \frac{c}{c_{\text{max}}} \times m_{\text{abs.}} \right\rfloor,
\]
where
$c_{\text{max}}$ is the maximum vertical (resp.\ horizontal) page extent,
$m_{\text{abs.}}$ is a training hyperparameter,
and $\lfloor \cdot \rfloor$ is the floor function.
The value $m_{\text{abs.}}$ act as a length normalization constant that defines the number of position embeddings to be learned.

Each discretized coordinate serves as an index for a learned embedding table.
In practice, there is a separate embedding table for each of the four bounding box coordinate.
The embeddings are then added to input tokens' embeddings.

\subsection{Relative Positions in \citet{xu_layoutlmv2_2022} and~\citet{Huang2022LayoutLMv3PF}}
\label{sec:t5_explanations}

\textsc{LayoutLMv2} \cite{xu_layoutlmv2_2022} and \textsc{v3} \cite{Huang2022LayoutLMv3PF} extend absolute positional information with relative information in the attention bias.
Instead of using the same discretization strategy for both, they rely on the bucketing strategy of \textsc{T5} \cite{raffel2020exploring} for relative distances:
rather than mapping distances linearly, they are assigned to buckets, with higher resolution for short distances and coarser resolution for long ones. 
This reflects the intuition that local order (e.g.\ adjacent words) requires precision, while very large distances can be captured approximately.

\subsection{Our Approach}
\label{sec:docpolarbert_bucketing}

\textsc{DocPolarBERT} departs from Cartesian coordinates and instead represents spatial relations in polar form.
For any pair of bounding box centers $(\vc_i,\vc_j)$, the distance $\rho(\vc_i,\vc_j)$ and angle $\theta(\vc_i,\vc_j)$ are discretized independently.

\paragraph{Angle.} For angles, we discretize $[-\pi, \pi ]$ into 8 buckets.
Each corresponds to a main direction (above, below, left, right, and four diagonals areas).
This reflects common document layouts, such as row/column alignments.

\paragraph{Distance.} Distances are partitioned into buckets based on empirical quantiles.
Unlike the scaling in the \textsc{LayoutLMv2/v3} bucketing strategy, the quantiles are computed from the distribution of distances \emph{within} each structured document.
Thus, the same value of relative distance can be assigned to a different bucket in two different documents.
This balances the representation by ensuring each bucket captures a comparable proportion of observed distances per document.
Instead of relying on a fixed maximum distance hyperparameter, our strategy adapts to the each document's distance distribution. 

By combining distance and orientation, the representation captures relationships crucial for forms and tables. 
Figure~\ref{fig:discretization_methods} illustrates the resulting scheme.

We compare this approach with alternative bucketing strategies in Section~\ref{sec:experiments}.

\begin{table}
\centering
\small
\begin{tabular}{lc}
\toprule
\textbf{Model} & \textbf{\# Parameters ($\downarrow$)}\\
\midrule
\multicolumn{2}{l}{\textbf{Document Understanding Models}}\\
\midrule
$\textsc{LayoutLMv2}_{\textrm{BASE}}$& 200M\\
$\textsc{LayoutMask}_{\textrm{BASE}}$& 182M\\
$\textsc{LayoutLM}_{\textrm{BASE}}$& 160M\\
$\textsc{LayoutLMv3}_{\textrm{BASE}}$& 133M\\
$\textsc{ReLayout}_{\textrm{BASE}}$& 125M\\
$\textsc{DocPolarBERT}_{\textrm{BASE}}$ (ours)& 125M\\
\midrule
\multicolumn{2}{l}{\textbf{Text-only Models}}\\
\midrule
$\textsc{RoBERTa}_{\textrm{BASE}}$& 125M\\
$\textsc{BERT}_{\textrm{BASE}}$& 110M\\
\bottomrule
\end{tabular}
\caption{Comparison of the number of parameters per model.}
\label{tab:parameters}
\end{table}

\section{Pre-training}

\paragraph{Neural architecture.}
We build a BERT-like encoder of 12 layers, with 12 heads per layer.
Queries, keys and embeddings are of dimension 768.
Beside our custom relative attention, other parts of the neural architecture follows the \textsc{LayoutLM} model \cite{xu_layoutlm_2020}, except that we \emph{do not introduce} absolute position information.

The input consists of a global 1D position embedding and a token embedding to capture semantic information.
Note that bounding box coordinates are used exclusively to compute distances and angles, which introduce spatial biases in the self-attention mechanism through a polar coordinate representation.
No vision features are required for this model.

Our model has $\simeq125$M parameters, making it ideal for efficient use in real-world applications as it is smaller than other layout models, see Table~\ref{tab:parameters}.

\paragraph{Training objectives.}
Following standard encoder-based models, we apply a masked language modeling loss with 30\% token masking.
Additionally, we experiment with the 1D local order prediction (1-LOP) loss \cite{jiang-etal-2025-relayout},
which predicts token positions in the linearized OCR output when 1D positional embeddings are masked (30\% of tokens).

\paragraph{Data.}
We initially sought to pre-train on the widely used \textsc{IIT-CDIP} dataset \cite{iit_cdip}, but its publicly available version lacks 2D layout annotations. 

To ensure reproducibility of our findings, and eliminate dependency on proprietary OCR systems, we opted for datasets in which text and layout annotations are publicly available: \textsc{OCR-IDL} \cite{ocr_idl_2023} and \textsc{DOCILE} \cite{simsa_docile_2023}. 
The \textsc{DOCILE} dataset consists of over 900k unlabeled public invoices, whereas \textsc{OCR-IDL} is a diverse collection of 26M documents (letters, reports, memos, news articles, etc.) sourced from the industry document library\footnote{\url{https://www.industrydocuments.ucsf.edu}} (IDL), the same repository used to construct \textsc{IIT-CDIP}. 
For computational efficiency and dataset balance, we sample 1.8M documents, evenly split between \textsc{DOCILE} and \textsc{OCR-IDL}.
While our final corpus is over six times smaller than \textsc{IIT-CDIP}, our results show that it is sufficiently large for model pre-training.

\paragraph{Training settings.}
We initialize from the weights of the pre-trained \textsc{RoBERTa} model \cite{liu2019robertarobustlyoptimizedbert} and use a minibatch size of 2048 documents.
We ran training for 20 epochs with a target learning rate of $1 \times 10^{-4}$ in a cosine scheduler \cite{loshchilov2017sgdr} with warmup on 5\% of the updates. 
Pre-training is done on NVIDIA Tesla V100 32GB GPUs.

\section{Experiments}\label{sec:experiments}

\subsection{Evaluation on Downstream Tasks}

\begin{table*}
\centering
\small
\resizebox{\textwidth}{!}{
\begin{tabular}{lccccc}
\toprule
 & \textbf{\textsc{Payslips}}& \textbf{$\textsc{DOCILE}$}&\textbf{\textsc{FUNSD}} &\textbf{\textsc{SROIE}} &\textbf{\textsc{CORD}}\\
\midrule
 Train / Val / Test sizes & 485/-/126 & 6,759/635/1,000 & 149/-/50 & 626/-/347  & 800/100/100\\
 \% of \textsc{O}  & 94.95& 89.46 & 0 & 83.82  & 0\\
 Document types & Pay Statements & Invoices & Forms & Receipts  & Receipts\\
 Entity types & Dates, Amounts & Text, Dates, Amounts & Text & Text, Dates, Amounts  & Text, Dates, Amounts\\
\bottomrule
\end{tabular}
}
\caption{Fine-tuning datasets description.}
\label{tab:ner_datasets}
\end{table*}
\begin{table*}
\centering
\small
\begin{tabular}{@{}lcccccc|c@{}}
\toprule
\textbf{Model / Pre-training data} & \textbf{Mod.} & \textbf{\textsc{FUNSD}} & \textbf{\textsc{SROIE}} & \textbf{\textsc{CORD}}  & \textbf{\textsc{Payslips}} &$\textbf{\textsc{DOCILE}}^*$ & \textbf{\textsc{Avg}}\\
\midrule
\multicolumn{8}{l}{\textbf{For reference only}}\\
\midrule
\rowcolor[gray]{0.9} $\textsc{LayoutLM}_{\textrm{BASE}}$ \cite{xu_layoutlm_2020} & T+L+I & \textit{79.27} & \textit{94.67} & -  & - &- & - \\
\rowcolor[gray]{0.9} $\textsc{LayoutLMv3}_{\textrm{BASE}}$\cite{Huang2022LayoutLMv3PF} & T+L+I & \textit{90.29}& - & \textit{96.56}  & - &- & - \\
\rowcolor[gray]{0.9} $\textsc{ReLayout}_{\textrm{BASE}}$  \cite{jiang-etal-2025-relayout} & T+L & \textit{84.64} & - & \textit{96.82}  & - &- & - \\
\rowcolor[gray]{0.9} $\textsc{LayoutMask}_{\textrm{BASE}}$  \cite{tu2023layoutmask} & T+L & \textit{92.91} & \textit{96.87} & \textit{96.99}  & - &- & - \\
\midrule
\multicolumn{8}{l}{\textbf{Baselines}} \\
\midrule
$\textsc{LayoutLM}_{\textrm{BASE}}$ (\textsc{IIT-CDIP}) & T+L  & \underline{\textit{78.66}}& \textit{94.38} & 95.66 & 62.31 &58.35 & 77.87\\
$\textsc{LayoutLM}_{\textrm{BASE}}$ (\textsc{DOCILE})  & T+L & 67.24 & 91.39 & 91.57  & 64.74 & 58.30 & 74.65 \\
 % \textsc{LayoutLMv3}$^*$ (\textsc{IIT-CDIP}& T+L& \underline{78.83}& 84.96& \textbf{96.97}& 62.39& 61.01&76.83\\
$\textsc{BROS}_{\textrm{BASE}}$ \cite{hong2022bros}& T+L & \textbf{\textit{83.05}} & \textit{96.28} & \textit{\textbf{96.50}}& 58.15 &57.09 & 78.21 \\
$\textsc{LiLT}_{\textrm{BASE}}$ \citep{wang-etal-2022-lilt} & T+L & 77.35 & 95.31 & \underline{\textit{96.07}}& 65.34 & 60.05 & 78.82 \\\midrule
\multicolumn{8}{l}{\textbf{\textsc{DocPolarBERT} (Ours)}}\\
\midrule
\textsc{DOCILE} & T+L & 57.28 & 94.02 & 91.78  & \textbf{76.85} &58.14 & 75.61 \\
+ \textsc{OCR-IDL} & T+L & 78.57 & \textbf{96.96} & 95.97  & 73.54 &\underline{59.73} & 80.95 \\
+ 1-LOP & T+L & 78.26 & \underline{96.53} & 95.15  & \underline{75.40} & \textbf{61.36} & \textbf{81.34} \\
\bottomrule
\end{tabular}
\caption{NER F1-score. Modalities are text (T), layout (L), and image (I). 
Results shown in \textit{italic} are from the original model papers. 
Bold and underline fonts emphasize best and second results for each dataset. }
\label{tab:results}
\end{table*}

\paragraph{Data.}
We evaluate our model on five datasets for named-entity recognition on documents.
FUNSD \cite{jaume-et-al-2019} is a dataset of 199 forms from the Truth Tobacco Industry Document (TTID) collection.
This collection is also the source of the IIT-CDIP \cite{iit_cdip} and the RVL-CDIP \cite{harley2015icdar} datasets.
SROIE and CORD v2 \cite{SROIE-2019, park2019cord} are datasets of respectively 973 and 1,000 annotated receipts.
\textsc{DOCILE} \cite{simsa_docile_2023} labeled subset with 6,680 annotated invoices.\footnote{Since the test set of DOCILE is not publicly available, we use the validation set for testing purposes.}
\textsc{Payslips} \cite{uthayasooriyar-etal-2025-training} is a dataset of 611 pages of financial documents (pay statements) from the insurance sector.
Dataset statistics are given in Table~\ref{tab:ner_datasets}.

\paragraph{Training settings.}
We fine-tune all models for 30 epochs with a minibatch size of 16 and a fixed learning rate of $5\times10^{-5}$, except for $\textsc{DOCILE}^*$ for which we use a learning rate of $1\times10^{-5}$.
To ensure robust evaluation, we report the average F1-score across 10 fine-tuning runs with different seeds, except for $\textsc{DOCILE}^*$, where we conduct only 4 runs to reduce computational costs. 

\paragraph{Results.}
We report results on these datasets for several baselines and for variants of our model in Table~\ref{tab:results}.
For reference, we also report results for models that are not directly comparable with \textsc{DocPolarBERT}, either because they are based on different modalities (vision features) or because they are not publicly available, hence we couldn't reproduce their results.

On all datasets except \textsc{FUNSD} and \textsc{CORD}, our model outperforms baselines.
We hypothesize \textsc{FUNSD}’s results are influenced by its origin from \textsc{IIT-CDIP}, used to pre-train all the other models.
For CORD, we observed that our model often confuses semantically similar or hierarchically related labels (e.g.\ subtypes within the ``menu'' category).
This is likely due to  limitations of polar-relative encoding in capturing fine-grained structures in dense regions with multiple mentions.

Interestingly, our approach largely outperforms baselines on \textsc{Payslips}, that contains mainly large tables of numbers.
This suggests our relative encoding strategy of layout structures better captures long-range dependencies,
particularly in cases where entity values must be correctly associated with corresponding column headers.

\subsection{Comparison with Vision-stripped \textsc{LayoutLMv3} Model}

% \begin{table*}[ht!]
% \centering
% \resizebox{\textwidth}{!}{
% \begin{tabular}{lccccc}
% \toprule
% \textbf{Model} & \makecell{\textbf{1-GPU Batch} \\ \textbf{(ms)}} & \makecell{\textbf{1-GPU Epoch} \\ \textbf{(ms)}} & \makecell{\textbf{4-GPU Batch} \\ \textbf{(ms)}} & \makecell{\textbf{4-GPU Epoch} \\ \textbf{(ms)}} & \makecell{\textbf{Speedup} \\ \textbf{(Batch / Epoch)}} \\
% \midrule
% \textsc{LayoutLM}         & \textbf{151.92} & 1873.42 & \textbf{76.43}  & 2265.64 & $\downarrow$49.7\% / \textcolor{red}{$\uparrow$21\%} \\
% \textsc{Layoutlmv3}$^*$       & \underline{202.58} & \textbf{1750.22} & \underline{88.19}  & \textbf{1768.07} & $\downarrow$56.5\% / \textcolor{red}{$\uparrow$1.0\%} \\
% \textsc{Lilt}             & 207.57 & \underline{1790.73} & 117.48 & 2100.89 & $\downarrow$43.4\% / \textcolor{red}{$\uparrow$17.3\%} \\
% \textsc{Bros}             & 345.97 & 2963.83 & 124.36 & 2318.74 & $\downarrow$64.1\% / $\downarrow$21.7\% \\
% \textsc{Docpolarbert}     & 413.59 & 3428.13 & 141.20 & \underline{1879.08} & \textbf{$\downarrow$65.9\%} / \textbf{$\downarrow$45.2\%} \\
% \bottomrule
% \end{tabular}
% }
% \caption{Average inference times across single- and multi-GPU settings. 
% Per-batch and per-epoch times are shown, along with the percentage change from 1-GPU to 4-GPU. Speedup is calculated separately for batch and epoch performance. 
% Tests were conducted on the \textsc{Payslips} dataset with a batch size of 16 and V100 GPUs.
% }
% \label{tab:inference-times}
% \end{table*}

\begin{table*}[ht!]
\centering
\small
\begin{tabular}{lccccc}
\toprule
\textbf{Model} &
\makecell{\textbf{Per-document} \\ \textbf{(1-GPU)}} &
\makecell{\textbf{Per-document} \\ \textbf{(4-GPU)}} &
\makecell{\textbf{Total dataset time} \\ \textbf{(1-GPU)}} &
\makecell{\textbf{Total dataset time} \\ \textbf{(4-GPU)}} &
\textbf{4-GPU Speedup} \\
\midrule
\textsc{LayoutLM}         & \textbf{9.5} & \textbf{4.8} & 1873.4 & 2265.6 & 0.83$\times$ \\
\textsc{Layoutlmv3}$^*$   & \underline{12.7} & \underline{5.5} & \textbf{1750.2} & \textbf{1768.1} & 0.99$\times$ \\
\textsc{Lilt}             & 13.0 & 7.3 & \underline{1790.7} & 2100.9 & 0.85$\times$ \\
\textsc{Bros}             & 21.6 & 7.8 & 2963.8 & 2318.7 & \underline{1.28}$\times$ \\
\textsc{Docpolarbert}     & 25.9 & 8.8 & 3428.1 & \underline{1879.1} & \textbf{1.82$\times$} \\
\bottomrule
\end{tabular}
\caption{
Inference efficiency comparison on the \textsc{PAYSLIPS} dataset.
Average per-document latency and total dataset processing time (in ms) across single- and multi-GPU setups.
GPUs are NVIDIA V100.
Speedup indicates the ratio of 1-GPU to 4-GPU total inference times.
Although our model is slower per document, its scales well over 4-GPUs.
}
\label{tab:inference-times}
\end{table*}

To assess the benefit of our polar-based representation in a vision-agnostic setting, we compare \textsc{DocPolarBERT} with a modified \textsc{LayoutLMv3} that excludes visual inputs, denoted $\textsc{LayoutLMv3}^*$. 
To this end, we pre-train the \textsc{LayoutLMv3}$^*$ model on the same corpus as ours, combining \textsc{DOCILE} and \textsc{OCR-IDL}, in order to isolate the effect of textual and layout information without interference from visual features.

As reported in Table~\ref{tab:layoutlmv3_vs_docpolarbert_mlm_idl}, \textsc{DocPolarBERT} consistently surpasses $\textsc{LayoutLMv3}^*$ across all downstream tasks, by a large margin. 
This indicates that encoding layout using relative polar coordinates better captures structural and semantic patterns from text and layout alone. 
In practical scenarios where visual data is unavailable, e.g.\ documents containing sensitive imagery in insurance or healthcare, models designed explicitly for textual and layout features, like \textsc{DocPolarBERT}, offer a clear advantage. 
In contrast, removing visual components from multimodal models like \textsc{LayoutLMv3} leads to degraded performance.

\subsection{Attention Analysis}

\begin{table}
\centering\small
\begin{tabular}{lcc}
\toprule
& $\textsc{LayoutLMv3}^*$  & \textsc{DocPolarBERT}  \\
\midrule
\textbf{\textsc{FUNSD}} &   66.79& \textbf{78.57} \\
\textbf{\textsc{SROIE}} &   89.83 &  \textbf{96.96} \\

 \textbf{\textsc{CORD}}  & 90.90 &\textbf{95.97}  \\
 \textbf{\textsc{PAYSLIPS}} &    44.44 &\textbf{73.54} \\
 $\textbf{\textsc{DOCILE}}^*$ &  59.66 & \textbf{59.73} \\
\midrule
 \textbf{\textsc{Avg}}& 70.32  &\textbf{80.95} \\
 \bottomrule
\end{tabular}
\caption{F1-score comparison of $\textsc{LayoutLMv3}^*$ and \textsc{DocPolarBERT}.
Both models are pre-trained on \textsc{OCR-IDL} + \textsc{DOCILE} with MLM loss only.}
\label{tab:layoutlmv3_vs_docpolarbert_mlm_idl}
\end{table}

\begin{figure}
    \centering
    \includegraphics[width=\linewidth]{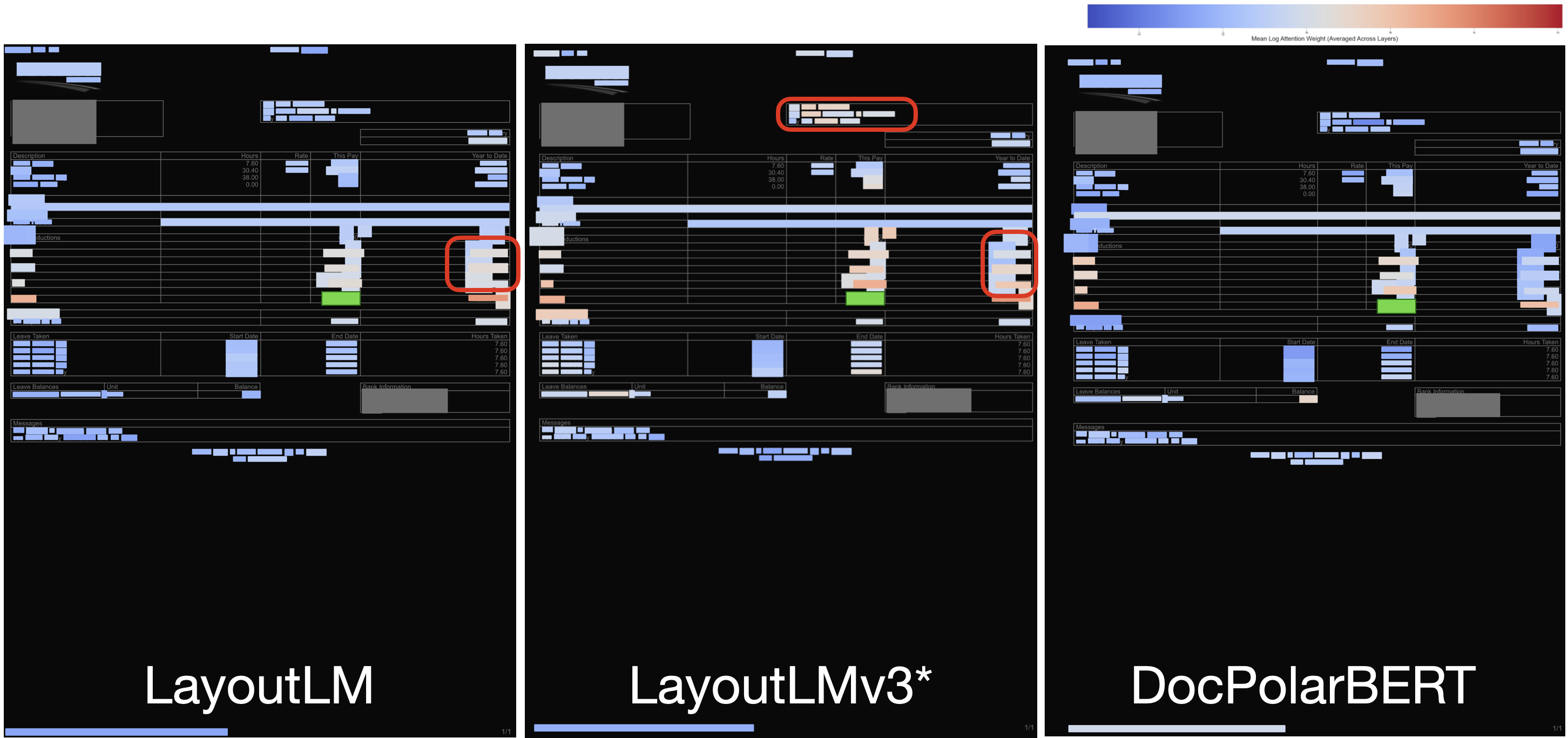}
    \caption{Heatmap of average attention (across all heads and layers) for a \texttt{NET\_PAY\_PER\_PERIOD} token (green box) on a \textsc{Payslips} document. 
    Blue-to-red values indicate low-to-high attention. 
    We compare \textsc{LayoutLM}, $\textsc{LayoutLMv3}^*$, and \textsc{DocPolarBERT}, highlighting with red bounding boxes the unwanted activations.}
    \label{fig:attention_amounts_correct}
\end{figure}
\begin{figure}
    \centering
    \includegraphics[width=\linewidth]{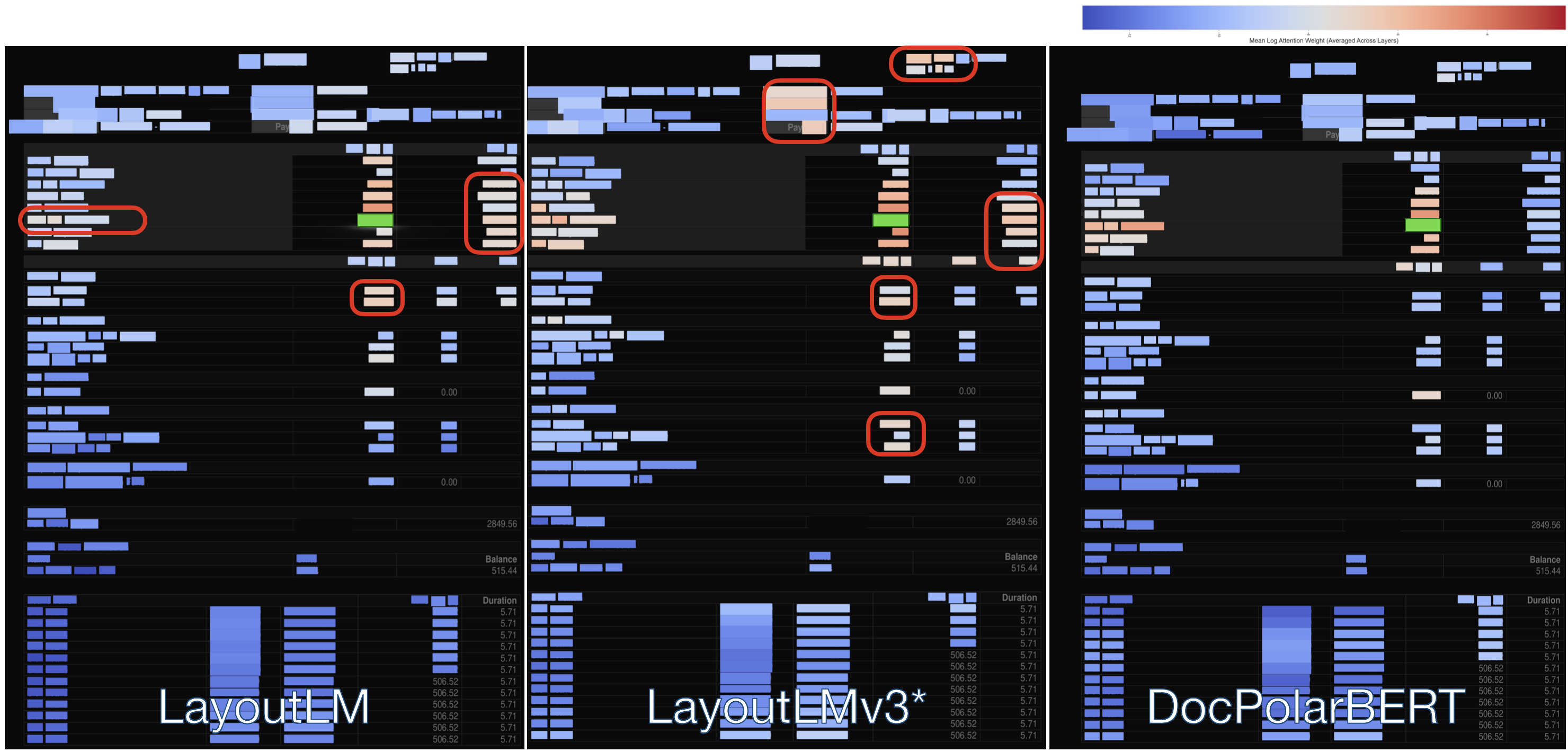}
    \caption{Heatmap of average attention for a \texttt{POST\_TAX\_DEDUCTIONS\_PER\_PERIOD} token (green box) on a \textsc{Payslips} document.
    We compare \textsc{LayoutLM}, $\textsc{LayoutLMv3}^*$, and \textsc{DocPolarBERT}, highlighting with red bounding boxes the unwanted activations.
    In this example, \textsc{LayoutLM} and \textsc{LayoutLMv3} misclassified the target amount, while \textsc{DocPolarBERT}  correctly identifies it.}
    \label{fig:attention_amounts_incorrect}
\end{figure}

To better understand how the proposed spatial bias affects model behavior, we manually examined attention distributions in fine-tuned models. 

First, we focus on samples from the \textsc{Payslips} dataset where target amounts were correctly classified.
For each document, we compare attention patterns of \textsc{LayoutLM}, \textsc{LayoutLMv3}$^*$, and \textsc{DocPolarBERT} with respect to the token corresponding to the predicted amount (Figure~\ref{fig:attention_amounts_correct}). 
In \textsc{LayoutLM} and \textsc{LayoutLMv3}$^*$, attention weights were frequently assigned to tokens without clear semantic or spatial relevance, such as unrelated text in distant parts of the document. 
By contrast, \textsc{DocPolarBERT} consistently concentrated on contextually and spatially coherent elements, including nearby numeric values and labels such as ``\texttt{Net pay}'', ``\texttt{Gross}'', or ``\texttt{Tax}'', often aligned horizontally or vertically with the target token.

We conducted a similar analysis on instances where only \textsc{DocPolarBERT} predicted the correct output while the other models failed (Figure~\ref{fig:attention_amounts_incorrect}). 
In these cases, \textsc{LayoutLM} and \textsc{LayoutLMv3}$^*$ also directed attention to irrelevant surrounding tokens or to distant header information, whereas \textsc{DocPolarBERT} maintained focus on semantically related and spatially aligned elements. 

Overall, encoding spatial relations in polar form seems to favor more coherent attention behavior, particularly in documents where layout strongly conveys meaning.

\subsection{Length Generalization}\label{sec:length_generalisation}
\begin{table*}[t]
\centering
\small
\begin{tabular}{lccc}
\toprule
& \small{\textsc{LayoutLM}} & \small{\textsc{LayoutLMv3}$^*$} & \small{\textsc{DocPolarBERT}} \\
\midrule
\textbf{\textsc{FUNSD}} & 54.02 & \textbf{70.47} & 62.10 \\
\textbf{\textsc{SROIE}} & 78.38 & \textbf{86.49} & 82.50 \\
\textbf{\textsc{CORD}} & 84.71 & \textbf{87.25} & 83.56 \\
\textbf{\textsc{PAYSLIPS}} & 77.80 & 74.22 & \textbf{85.10} \\
$\textbf{\textsc{DOCILE}}^*$ & 60.17 & 54.45 & \textbf{61.04} \\
\midrule
\textbf{\textsc{Avg}} & 71.02 & 74.58 & \textbf{74.86} \\
\bottomrule
\end{tabular}

\caption{F1-scores of \textsc{LayoutLM}, $\textsc{LayoutLMv3}^*$, and \textsc{DocPolarBERT} on the sequence-length generalization task. Models are trained on the shortest documents and tested on the longest ones. All are pre-trained on \textsc{DOCILE} using masked language modeling only.}
\label{tab:new_splits}
\end{table*}

Length generalization is a common setting in Transformer evaluation \cite[][\emph{inter alia}]{bhattamishra-etal-2020-ability,anil2022,wang-etal-2024-length} and structured prediction \cite{herzig-berant-2021-span,yao-koller-2022-structural,wu-etal-2023-recogs}.
We propose to adapt this setting to document understanding.
Each dataset is sorted by token length,
and we split the datasets between train and test sets so that they contains the same number of documents in each split as the original dataset.
The training set includes the shortest documents, while the test set contains the longest ones.

We evaluate \textsc{LayoutLM}, \textsc{LayoutLMv3}$^*$, and \textsc{DocPolarBERT}, pre-trained with masked language modeling on \textsc{DOCILE}.
The results are shown in Table~\ref{tab:new_splits}.
\textsc{DocPolarBERT} achieves the highest average F1-score, slightly surpassing both \textsc{LayoutLM} and \textsc{LayoutLMv3}$^*$. 

However, performance differences across datasets reveal more nuanced patterns. 
On \textsc{FUNSD}, \textsc{DocPolarBERT} underperforms compared to \textsc{LayoutLMv3}$^*$, with similar behavior on \textsc{SROIE} and \textsc{CORD}. 
This difference can be attributed to the relatively simple spatial layouts of these datasets, where most information aligns along clear horizontal and vertical axes. 
In such cases, Cartesian coordinates offer an adequate way to capture spatial relationships, while polar coordinate system may add over-complex geometric abstraction.

In contrast, \textsc{DocPolarBERT} performs best on \textsc{Payslips} and $\textsc{DOCILE}^*$, which contain highly structured tables with clear header–value pairings. 
In these documents, the polar encoding captures relational patterns soundly, even when document size or layout changes.
This suggests that polar-relative encoding introduces a strong structural bias beneficial for layout-driven understanding of documents with long tables.

\subsection{Efficiency analysis}

\textsc{DocPolarBERT} shows a noticeable slowdown during inference compared to baseline architectures. 
As shown in Table~\ref{tab:inference-times}, per-document inference times are higher, primarily due to the size and structure of the relative bias matrices in our attention mechanism.

Despite the higher per-document cost, \textsc{DocPolarBERT} demonstrates superior scaling efficiency on multi-GPU setups. 
Table~\ref{tab:inference-times} shows substantial reductions in  inference times over the whole dataset when moving from 1-GPU to 4-GPUs.\footnote{4-GPU epoch times can exceed 1-GPU times due to fixed multi-GPU overheads, such as inter-GPU communication and synchronization, which may outweigh parallelization gains for already optimized single-GPU tasks.}

\section{Ablation Study}

\subsection{Absolute Positional Embeddings}

\begin{table}
\centering\small
\begin{tabular}{lcc}
\toprule
& w 2D-Pos  & w/o 2D-Pos  \\
\midrule
\textbf{\textsc{FUNSD}} & 76.22 & \textbf{78.26} \\
\textbf{\textsc{SROIE}} & 96.52  & \textbf{96.53} \\

 \textbf{\textsc{CORD}}  & 94.63 &\textbf{95.15}  \\
 \textbf{\textsc{Payslips}} & 72.52  &\textbf{75.40} \\
 $\textbf{\textsc{DOCILE}}^*$ & 61.27 & \textbf{61.36} \\
\midrule
 \textbf{\textsc{Avg}}& 80.23 &\textbf{81.34} \\
 \bottomrule
\end{tabular}
\caption{\textsc{DocPolarBERT} F1-score comparison with and without absolute 2D positional embeddings.
}
\label{tab:appendix_2D_pos_comparison}
\end{table}

Contrary to previous work, our model does not include absolute 2D positional embeddings.
We therefore conduct a study to analyse the impact of absolute positional embeddings in \textsc{DocPolarBert}.

Table~\ref{tab:appendix_2D_pos_comparison} compares NER results for \textsc{DocPolarBERT} with and without such embeddings.
We observe that absolute 2D positional embeddings, that are always provided in baselines, deteriorates downstream task results.

\subsection{Discretization Strategies}

\begin{table*}[t]
\centering\small
\begin{tabular}{lcccc@{\hspace{1cm}}cccc}
\toprule
Bucketing method &\multicolumn{4}{c}{\textsc{LayoutLMv2/v3}} & \multicolumn{4}{c}{Quantile-based}\\
\midrule
Nb. of buckets& 32 & 16&     8&4&32&16&8&4\\
\midrule
\textbf{\textsc{FUNSD}} & 79.06& 78.81&     78.51&\textbf{80.01}&79.44&76.71&79.90&78.26\\
\textbf{\textsc{SROIE}} & 97.42& 97.69&     97.31&97.51&\textbf{97.74}&97.14&97.68&96.53\\
\textbf{\textsc{CORD}} & 96.94& \textbf{97.38}&     96.46&97.23&96.77&96.92&96.78&95.15\\
\textbf{\textsc{PAYSLIPS}} & 73.76& 73.40&     73.37&73.53&73.66&72.81&74.27&\textbf{75.40}\\
$\textbf{\textsc{DOCILE}}^*$ & 57.87& 58.82&     57.82&55.00&58.31&57.83&56.06&\textbf{61.36}\\
\midrule
\textbf{\textsc{Avg}} & 81.01& 81.22&     80.70&80.66&81.18&80.28&80.94&\textbf{81.34}\\
\bottomrule
\end{tabular}
\caption{NER F1-score of \textsc{DocPolarBERT} with the bucketing method from \textsc{LayoutLMv2/v3} vs our quantile-based strategy applied to relative distances. 
Each variant is pre-trianed on \textsc{OCR-IDL}+\textsc{DOCILE} with MLM and 1-LOP.}
\label{tab:buckets}
\end{table*}

As discussed in Section~\ref{sec:discretization},
previous models use a different distretization method for encoding relative position information.
We therefore conduct experiments to evaluate their impact.

In the following, we compare variant of \textsc{DocPolarBERT} where the distance coordinate is discretized:
\begin{itemize}
    \item Using a similar strategy as in \textsc{LayoutLMv2} and \textsc{v3}, with varying number of bins (4 to 32), see Section~\ref{sec:t5_explanations};
    \item Using our quantile-based strategy, but with different number of bins (4 to 32), see Section~\ref{sec:docpolarbert_bucketing}.
\end{itemize}
Results are given in Table~\ref{tab:buckets}.

Overall, results remain close across strategies, with less than two points variation in average F1. 
This shows that \textsc{DocPolarBERT} is generally robust to the exact choice of bucketing.

Nonetheless, some dataset-specific patterns are visible when considering the number of buckets. 
On \textsc{FUNSD}, using fewer buckets appears advantageous, with the T5 4-bucket variant reaching the highest score ($80.01$). 
In contrast, \textsc{SROIE} benefits from finer granularity, where the quantile variant with 32 buckets achieve the best results ($97.74$). 
\textsc{CORD} shows a consistent preference for 16 buckets across both strategies. 
For \textsc{PAYSLIPS} and $\textsc{DOCILE}^*$, quantile-based schemes perform slightly better, with the 4-quantile variant reaching the highest scores, which suggests that adapting buckets to the empirical distribution can be advantageous for tabular or invoice-like layouts. 

On average, the 4-quantile approach achieves the best performance ($81.34$), albeit by a narrow margin over the best T5 variant ($81.22$). 
Therefore, while the overall effect of bucket choice is limited, adapting bucket counts to dataset characteristics can offer small but consistent benefits. 

\section{Conclusion}

We propose a novel relative positional encoding of layout structure based on the polar coordinate system.
This approach allows the model to directly encode semantically crucial information (e.g.\ table headers are at the top row, no matter how far it is).
Experimentally, our approach outperforms comparable models.

\section{Limitations}

Our proposed document encoder model requires text and layout annotations from an OCR model, which is different from end-to-end systems that directly process raw documents.
Note that although running an OCR can be a costly step, it is nevertheless often done in practical settings,
where the data is preprocessed before being archived.

We evaluate our approach only on named-entity recognition, which is the most common benchmark for layout models, but it may not cover all use cases.

Finally, our pre-training data consists of a mix of \textsc{OCR-IDL} and \textsc{DOCILE}, totaling 1.8M documents, whereas many document understanding models leverage \textsc{IIT-CDIP}, which contains 11M documents.
We opted for datasets with openly available text and layout annotations to ensure reproducibility.
This choice facilitates fair comparison and avoids the reproducibility issues common in document understanding research, where OCR outputs are often proprietary and can heavily affect downstream performance.

\section*{Acknowledgement}

This work was performed using HPC resources from GENCI-IDRIS (Grant 2024-AD011015001).

\bibliography{custom}

\end{document}